\documentclass[journal]{IEEEtran}
\usepackage{bookmark}
\usepackage{ifpdf}
\usepackage{cite}
\usepackage[pdftex]{graphicx}
\usepackage{amsmath}
\usepackage{algorithmic}
\usepackage{array}
\usepackage{float}
\usepackage{dblfloatfix}
\usepackage[dvipsnames]{xcolor}
\usepackage{graphicx}
\usepackage{subcaption}
\usepackage{multirow}
\usepackage[switch]{lineno}
\usepackage{lipsum}
\usepackage{scalerel}
\usepackage{tikz}
\usetikzlibrary{svg.path}

\definecolor{orcidlogocol}{HTML}{A6CE39}
\tikzset{
	orcidlogo/.pic={
		\fill[orcidlogocol] svg{M256,128c0,70.7-57.3,128-128,128C57.3,256,0,198.7,0,128C0,57.3,57.3,0,128,0C198.7,0,256,57.3,256,128z};
		\fill[white] svg{M86.3,186.2H70.9V79.1h15.4v48.4V186.2z}
		svg{M108.9,79.1h41.6c39.6,0,57,28.3,57,53.6c0,27.5-21.5,53.6-56.8,53.6h-41.8V79.1z M124.3,172.4h24.5c34.9,0,42.9-26.5,42.9-39.7c0-21.5-13.7-39.7-43.7-39.7h-23.7V172.4z}
		svg{M88.7,56.8c0,5.5-4.5,10.1-10.1,10.1c-5.6,0-10.1-4.6-10.1-10.1c0-5.6,4.5-10.1,10.1-10.1C84.2,46.7,88.7,51.3,88.7,56.8z};
	}
}

\newcommand\orcidicon[1]{\href{https://orcid.org/#1}{\mbox{\scalerel*{
				\begin{tikzpicture}[yscale=-1,transform shape]
				\pic{orcidlogo};
				\end{tikzpicture}
			}{|}}}}

\DeclareMathOperator*{\argmin}{argmin}
\usepackage{hyperref} 

\begin{document}
{\textcopyright}2020 IEEE. Personal use of this material is permitted. Permission from IEEE must be obtained for all other uses, in any current or future media, including reprinting/republishing this material for advertising or promotional purposes, creating new collective works, for resale or redistribution to servers or lists, or reuse of any copyrighted component of this work in other works. This paper is a preprint version.

\newpage

\title{Stereo Visual Inertial Pose Estimation Based on Feedforward-Feedback Loops}
\author{Shengyang~Chen \orcidicon{0000-0003-1506-0615}\,,
		Chih-Yung~Wen \orcidicon{0000-0002-1181-8786}\,,
        Yajing~Zou \orcidicon{0000-0002-5476-3826}\,
        and~Wu~Chen \orcidicon{0000-0002-1787-5191}\,
\thanks{All authors are with The Hong Kong Polytechnic University, Hong Kong}
\thanks{S.~Chen and C.-Y.~Wen are with the Department of Mechanical Engineering and Interdisciplinary Division of Aeronautical and Aviation Engineering e-mail: shengyang.chen@connect.polyu.hk; cywen@polyu.edu.hk}
\thanks{Y.~Zou and W.~Chen are with the Department of Land Surveying and Geo-Informatics e-mail: rick.zou@connect.polyu.hk; lswuchen@polyu.edu.hk}
\thanks{Manuscript received June 30, 2020; revised Xxx xx, 2020.}}

\maketitle

\begin{abstract}
In this paper, we present a novel stereo visual inertial pose estimation method. Compared to the widely used filter-based or optimization-based approaches, the pose estimation process is modeled as a control system. Designed feedback or feedforward loops are introduced to achieve the stable control of the system, which include a gradient decreased feedback loop, a roll-pitch feed forward loop and a bias estimation feedback loop. This system, named FLVIS (Feedforward-feedback Loop-based Visual Inertial System), is evaluated on the popular EuRoc MAV dataset. FLVIS achieves high accuracy and robustness with respect to other state-of-the-art visual SLAM approaches. The system has also been implemented and tested on a UAV platform. The source code of this research is public to the research community.
\end{abstract}


%
\IEEEpeerreviewmaketitle

\section{Introduction}
\IEEEPARstart{F}{using} the measurement of IMU and an extra camera extends the monocular vSLAM system to a stereo visual inertial system. The advantages of such system can be categorized into three aspects:
\begin{itemize}
	\item Robustness: The pose between consecutive visual frames can be estimated by IMU. In addition, when the visual tracking is lost, IMU can maintain the pose output within a short period.
	\item Accuracy: More measurements are fused in the pose estimation process, leading to better accuracy.
	\item Scale-consistency: The depth information can be extracted directly from stereo images without any motion. The system scale is consistent, and the initialization can be achieved in one shot.
\end{itemize}
\begin{figure}[hbt!]
	\centering
	\begin{subfigure}{1.0\linewidth}
		\centering
		\includegraphics[width=0.95\linewidth]{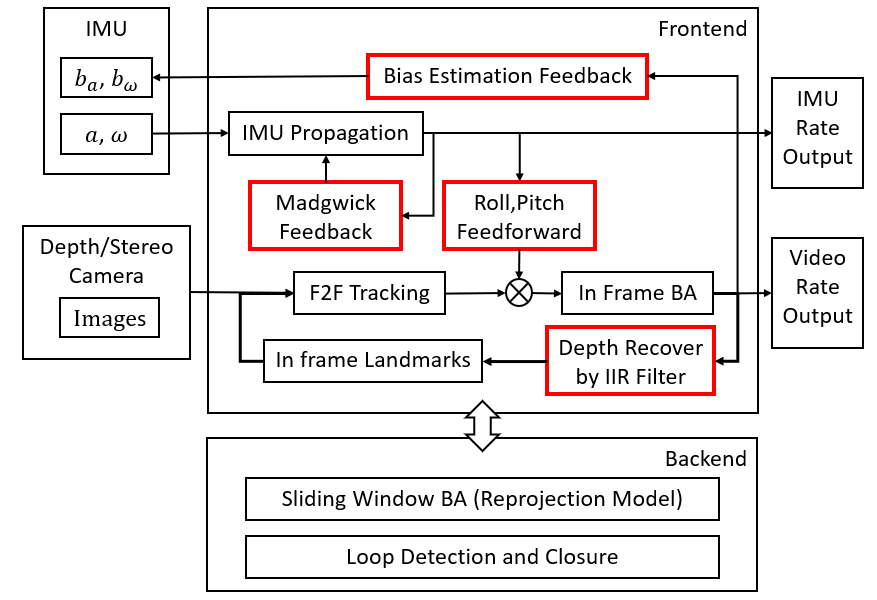}
		\caption{Simplified system workflow}
	\end{subfigure}\\
	\begin{subfigure}{1.0\linewidth}
		\centering
		\includegraphics[width=0.95\linewidth]{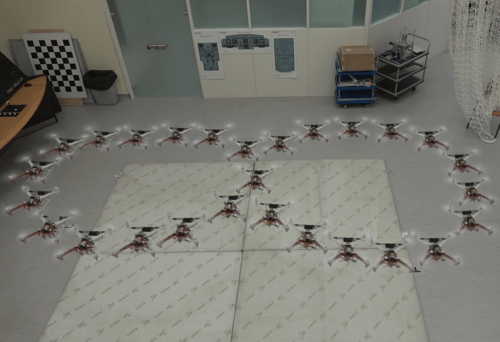}
		\caption{UAV flying figure-8 trajectory with FLVIS (composite image)}
	\end{subfigure}
	\caption{Simplified system workflow and the UAV test}
	\label{fig:sim_arch_and_euroc_result}
\end{figure}

However, the above advantage comes with a price. The visual information handled by the stereo visual inertial system is twice as large as that of the monocular system. In addition, a typical IMU sensor delivers a data rate on the order of one hundred Hz. The modern feature-based viSLAM adopts either the filter-based framework or the optimization-based framework to fuse all of these data.

In the filter-based framework, the pose and the landmark are included in system states. The IMU inputs propagate the pose states and the relevant parts in the covariance matrix. The propagation result acts as the prediction for the next visual input. When the next visual input arrives, it updates all states though the measurement model and extends the landmark states with the new extracted features. 

In the graph optimization-based framework, the camera poses and landmarks are presented as vertices. A reprojection edge and a preintegration edge are defined by two adjacent vertices in the visual measurement and in the propagated IMU measurement, respectively. The pose estimation is carried out by minimizing the objection functions of two types of edges.

Both frameworks described above are based on finding the most appropriate prediction to satisfy every single measurement (visual and IMU). As the amount of input data increases, more computational power is required. For a robotic system, especially a payload sensitive system, such as UAV, the perception algorithm may run in an embedded computer. Thus, there is a conflict between the limited computational resources and the real-time processing requirement.

The motivation of this work is to address the accurate VI fusion problem for resource limited UAV navigation. We present a feedforward-feedback loop based visual inertial system (or FLVIS). This framework utilizes the advantages of contemporary frameworks along with several independent models to decouple and reassemble the stereo visual inertial fusing process. The ideal measurement model assumption is made, which means that the errors induced by other settings, such as inaccurate camera calibration, were neglected. The accuracy of the ego-motion estimation only depends on (1) the accuracy of the landmark position and (2) the quality of the initial guess.

Therefore, a better initial guess from IMU and well maintained positions of landmarks are closely related to the accuracy of the system. The simplified system architecture is shown in Figure 1. The motion core is based on the standard IMU propagation model, in which a one-step gradient-based filter (Madgwick filter) is added. This filter complements the accelerometer and the gyroscope to achieve an accurate and robust orientation estimation. Then, the IMU states are fed forward into the classical vision estimation pipeline as the correction of the initial guess for the in-frame bundle adjustment (BA). After the in-frame BA, the biases of the accelerometer and the gyroscope are estimated and fed back to the IMU propagation model. Meanwhile, the depth information of every frame is extracted, and the positions of landmarks are updated through an infinite impulse response (IIR) filter.

Different from the conventional framework which uses the keyframe technology in the frontend and adopts frame to keyframe tracking to restrain the accumulation of drift \cite{leutenegger2013keyframe}, in the current framework, the frontend only forwards the keyframe message to the backend in order to speed up processing. No keyframe information is maintained in the frontend. The backend of the system is a classic sliding window optimizer-based on the reprojection model, which optimizes all keyframe measurements and passes the correction to the frontend. The loop closure is also implemented in this system to achieve correction in large scenarios.

Finally, the current method is verified by the public dataset and in the real-world environment. The evaluation results show that the accuracy of our system is comparable to the state-of-art methods. Additionally, our method is applied to a resource limited UAV system. The UAV is demonstrated to achieve on-board perception and control in an indoor environment (Figure \ref{fig:sim_arch_and_euroc_result}). In summary, the contributions of this work include:

\begin{itemize}
	\item Modeling the stereo visual inertial pose estimation as a control perspective.
	\item Applying a feedback/feedforward loop to achieve the sensor fusion.
	\item Adopting an Infinite impulse response (IIR) filter for landmark updating.
	\item Implementing an open-source stereo viSLAM for the research community\footnote{https://github.com/Ttoto/FLVIS}.
\end{itemize}

\section{Related Work}
Various works regarding visual SLAM have been conducted in the past decades. In this section, the related works are reviewed according to the following aspects:
\subsection{Notable vSLAM Work}

The SLAM problem, in its early age, is modeled on the basis of the Markov assumption and the sensor measurement models \cite{thrun2002probabilistic}. Typically, such a problem can be solved by using the extended Kalman filter (EKF). MonoSLAM by Davison \cite{davison2003real} is the most notable work of this kind. In MonoSLAM, the image patches, serving as landmark features, are measured from one frame to another. The pose estimation and the map recovery are carried out by interactive evolution of the probability densities over the feature depth and the camera pose in the EKF framework.

With the development of SLAM theory, the visual pose estimation is later modeled as pose graph optimization on a manifold \cite{grisetti2010hierarchical} \cite{mouragnon2006real}. As mentioned earlier in the Introduction section, in the pose graph optimization, camera poses and landmarks are presented as vertices, and the corresponding measurements are presented as edges. Based on the gradient distributions, the optimizer adjusts the vertices in their neighborhood region to minimize the cost function of edges. The widely used g2o \cite{kummerle2011g} is one such graph optimization framework.

PTAM (parallel tracking and mapping) \cite{klein2007parallel} is the next milestone of vSLAM study. As the name implies, the system splits the vSLAM problem into two separate components, tracking and mapping, and handles them in parallel on the multicore computer. The tracking thread (as the frontend) estimates the camera pose according to the camera frame rate, while the mapping thread (as the backend) performs BA on keyframes at a reduced frame rate. Such a frontend/backend design is the prototype of most modern vSLAM systems.

Generally, the optimization models of vSLAM can be further categorized into two groups: the feature-based method and the direct method (also known as the dense method). The first group of methods uses the sparse features. The object function is created on the basis of reprojection errors of these features. The most notable work of this kind is ORB-SLAM. Compared to its ancestors, ORB-SLAM includes many well-developed components: ORB features \cite{rublee2011orb}, dictionary-based loop detection \cite{GalvezTRO12}, scale-aware loop closing \cite{strasdat2010scale}, covisibility-based optimization \cite{strasdat2011double} and keyframe management strategy. This system has demonstrated its high accuracy and reliability in many different scenarios.

The second group uses the information from the entire image, and the objection function is based on the photometric error. The greatest advantage of using the direct method is that the depth is recovered during the process and can output a dense map directly. The dense map, compared to the sparse map, provides more details and can be easily used in navigation applications. DTAM (dense tracking and mapping in real-time) \cite{newcombe2011dtam} is the most notable work of this kind.

In contemporary times, the state-of-the-art vSLAM system bridges the gap between feature-based and direct methods and utilizes ideas from both of them. The SVO (semidirect visual odometry) \cite{forster2014svo} uses sparse features and the direct method to reach an extremely high processing speed and is used together with the reprojection error-based BA to achieve good accuracy. In DSO (direct sparse odometry) \cite{engel2017direct} and LSD-SLAM (large-scale direct monocular SLAM) \cite{engel2014lsd}, a gradient-based semi-dense feature is adopted and accordingly reduces the computational load as compared to the dense system.

\subsection{Visual Inertial Fusion in viSLAM study}
Visual inertial SLAM (viSLAM), as a branch of vSLAM, is focused on fusing the inertial measurement to the vSLAM system to increase its accuracy and robustness. Early attempts at applying VI fusion include the VI works of Diel et al. \cite{diel2005epipolar} and Oskiper et al. \cite{oskiper2007visual}. During these early stages of studies, IMU served as an independent orientation sensor and was fused with the vision estimation result through filters. In other words, the visual camera and IMU present their own perception results. These two results are then fused by a filter. Such a system is called a loosely coupled system.

Compared with the loosely coupled system, the tightly coupled system includes both the IMU states and visual measurement in their estimation framework, which are innovated together. In MSCKF \cite{mourikis2007multi} and ROVIO \cite{bloesch2015robust}, the nominal state, together with its covariance matrix, is updated in the IMU propagation process and innovated when a visual measurement arrives. The greatest advantage of such tightly coupled systems is that the egomotion between camera frames can be recovered with the IMU propagation model, therefore achieving nearly real-time egomotion estimation and yielding the explosion in the use of VI systems in aggressive robot applications such as UAV navigation.

The next evolution of VI fusion is the ”preintegration” theory. Since the optimization-based system has become popular, the results of preintegration can be integrated into the graph-based optimizer. Among them, OKVIS \cite{leutenegger2013keyframe} and VINSMono \cite{leutenegger2013keyframe} are the benchmarks for in viSLAM studies.

\subsection{Monocular or Stereo/Depth vSLAM} 
In the monocular vSLAM system, an unavoidable challenge is the scale recovery. Among all monocular vSLAM systems, the scale is handled by an independent process with several alternative solutions, e.g., IMU integrated SFM \cite{qin2018vins}, predefined object pattern \cite{frost2018recovering} \cite{pfrommer2019tagslam} and geometric perspective estimation \cite{wang2018monocular}. Because all of these solutions are either based on the preliminary message of the environment or require the camera motion, the application scenarios are limited.

Apart from monocular cameras, nowadays, there are many depth cameras available with reasonable prices. These depth sensors can be categorized into three types: stereo-based \cite{okutomi1993multiple}, ToF-based \cite{foix2011lock}, and structured light \cite{geng2011structured}-based cameras. Though different in their sensing modes, all of these are depth cameras and can sense depth information of a scene immediately without the need for motion. This advantageous feature promotes the studies of depth vSLAM.

Since the depth information can be extracted from every frame, depth vSLAM can be modeled as a point cloud registration problem. The egomotion is estimated by an iterative closest point (ICP) process \cite{besl1992method}. Kinect-fusion \cite{izadi2011kinectfusion} and DVO \cite{kerl2015dense} are the most notable works of this kind. Other depth vSLAM systems can be regarded as the extensions of their monocular versions, such as VINS-Fusion \cite{qin2019general} of VINS-mono, ORBSLAM2 \cite{mur2017orb} of ORB-SLAM and SVO 2.0 \cite{forster2016svo} of SVO. In these works, the measurement of another camera and the stereo constraints were added into the optimizer frameworks. As more information is integrated, these depth vSLAM methods are more robust than their monocular counterparts. However, depth vSLAM systems demand the price of greater computational power.

\section{System Overview and Notation}
\begin{figure}[H]
	\centering
	\includegraphics[width=0.8\linewidth]{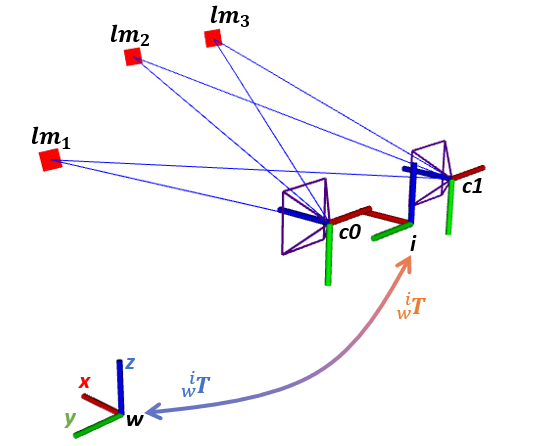}
	\caption{Coordinates of stereo visual inertial system}
	\label{fig:svio_coor}
\end{figure}
\begin{figure*}[!b]
	\centering
	\includegraphics[width=1.0\linewidth]{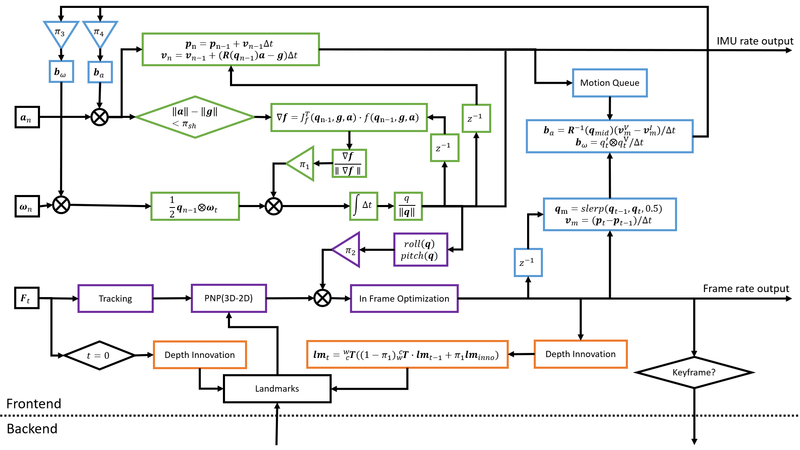}
	\caption{Frontend Workflow}
	\label{fig:feworkflow}
\end{figure*}
A stereo visual inertial system, as shown in Figure \ref{fig:svio_coor}, consists of an IMU module and two independent cameras. The coordinate system on the upper right part of Figure 2 is attached to the IMU center and designated as the IMU frame ($\boldsymbol{i}$). The corresponding optical center coordinates of two cameras are camera0 frame ($\boldsymbol{c}_0$) and camera1 frame ($\boldsymbol{c}_1$), respectively. The world frame is defined as a local east, north, and up (ENU) Cartesian coordinate system (shown in the lower left corner of Figure \ref{fig:svio_coor}), which means that the direction of gravity is opposite the z-axis. The transformation $\boldsymbol{T}$ between these frames is represented by the manifold on the special Euclidean group (SE(3)). For example, the symbol $_i^w\boldsymbol{T}$ refers to the transformation from the IMU frame to the world frame:

\begin{equation}
_i^w\boldsymbol{T} = \begin{pmatrix} _{i}^{w}\boldsymbol{R}_{3\times 3} & _{i}^{w}\boldsymbol{p}_{3\times 1}\\ 
\boldsymbol{0}_{1\times 3} & 1 \end{pmatrix} = (_i^w\boldsymbol{q},_i^w\boldsymbol{p}) \in SE(3)
\label{equ:notation_transformation}
\end{equation}

In Equation \ref{equ:notation_transformation}, $ _{i}^{w}\boldsymbol{R}_{3\times 3}$ refers to the rotation matrix from the IMU frame to the world frame. It can also be parameterized using the quaternion $_i^w\boldsymbol{q}$. The $_{i}^{w}\boldsymbol{p}_{3\times 1}=(p_x,p_y,p_z)^T$ refer to the position displacement (or translation) from the IMU frame to the world frame.

Through the stereo calibration and the visual inertial calibration, we can obtain the extrinsic parameters of the system, which include the installation geometry of the stereo camera $_{c_1}^{c_0}\boldsymbol{T}$ and the visual inertial installation geometry $_{c_0}^{i}\boldsymbol{T}$. As these extrinsic parameters are fixed, the sensor frame is associated with the IMU frame. Consequently, the egomotion can be described using: rotation $_i^w\boldsymbol{q}$, position $_i^w\boldsymbol{p}$ and velocity $_i^w\boldsymbol{v}$. Notably, these states are defined in the word frame. For simplicity, subscripts and superscripts are neglected, and $\boldsymbol{q}$ $\boldsymbol{p}$ $\boldsymbol{v}$ are used hereinafter.

The state of the system also contians the IMU related states and local landmarks. The IMU states include the IMU data ($^i\boldsymbol{a}$, $^i\boldsymbol{\omega}$) and the estimated bias ($^i\boldsymbol{b}_a$, $^i\boldsymbol{b}_\omega$) in the IMU frame. For simplification, $\boldsymbol{a}$, $\boldsymbol{\omega}$, $\boldsymbol{b}_a$ and $\boldsymbol{b}_\omega$ are used to present them. A feature captured in frame $\boldsymbol{c}_0$, after obtaining its depth information from the stereo image, becomes the landmark $\boldsymbol{lm}_i=(x_i,y_i,z_i)^T$ and is added into the states as the local map information. The landmarks will be tracked in consecutive frames and updated by the new measurement through an IIR filter. The landmarks will disappear and fade out from the state when they are no longer inside the image or when the tracking has failed.

In summary, the full state of the system can be represented by $\boldsymbol{x}=(\boldsymbol{q},\boldsymbol{p},\boldsymbol{v}, \begingroup \color{magenta} \boldsymbol{a} \endgroup, \begingroup \color{magenta} \boldsymbol{\omega} \endgroup ,\boldsymbol{b}_a,\boldsymbol{b}_\omega, \begingroup \color{cyan} \boldsymbol{lm}_i \endgroup , \begingroup \color{cyan} \boldsymbol{lm}_{i+1} \endgroup, \begingroup \color{cyan} ... \endgroup, \begingroup \color{cyan} \boldsymbol{lm}_{i+n} \endgroup)^T$. The state will be updated when either IMU or visual measurement is available. Notably, the state parameters are colored, where those in \begingroup \color{magenta} magenta \endgroup are only related to the inertial measurement and those in \begingroup \color{cyan} cyan \endgroup are only related to the visual information. The other states are the coupled states, in which the $\boldsymbol{q}$,$\boldsymbol{p}$ and $\boldsymbol{v}$ can be derived from both sensors, and the biases $\boldsymbol{b}_a$ and $\boldsymbol{b}_\omega$ engaged in the IMU propagation process can be estimated from the visual measurement.

\section{Frontend}
The frontend can be roughly divided into four modules: three as feedback/feedforward and one as depth information. These modules are colored in Figure 3 as IMU propagation with Madgwick feedback loop \textcolor{OliveGreen}{(in green)}, visual estimation with orientation feedforward \textcolor{RoyalPurple}{(in purple)}, interframe bias estimation feedback \textcolor{cyan}{(in blue)} and depth recovery and IIR filter \textcolor{BurntOrange}{(in orange)}. These modules will be introduced in sequence as follows:

\subsection{IMU Propagation with Madgwick Feedback Loop }
\label{ssec:imumadwick}
\subsubsection{IMU Sensor Model}

The MEMS IMU consists of a 3-axis accelerometer and a 3-axis gyroscope. The measured angular velocity and acceleration can be described by the following models:

\begin{align}
&\boldsymbol{\omega}_{m}=\boldsymbol{\omega}_{real} + \boldsymbol{b}_{\omega}+\boldsymbol{n}_{\omega}\\
&\boldsymbol{a}_{m} = \boldsymbol{a}_{real}+\boldsymbol{b}_{a}+\boldsymbol{n}_{a}
\end{align}
Where $\boldsymbol{n}_{\omega}$ and $\boldsymbol{n}_{a}$ refer to the intrinsic noises of the sensor which follow the Gaussian distributions:
\begin{equation}
\boldsymbol{n}_{\omega} \sim \mathcal{N}(\boldsymbol{0},\boldsymbol{\sigma}_{\omega}^2); \boldsymbol{n}_a \sim \mathcal{N}(\boldsymbol{0},\boldsymbol{\sigma}_a^2)
\end{equation} 
The biases ($\boldsymbol{\omega}_{b}$ and $\boldsymbol{a}_{b}$) are affected by the temperature and change over time. The time derivatives of these biases also follow the Gaussian distributions:
\begin{equation}
\boldsymbol{\dot{b}}_\omega \sim \mathcal{N}(\boldsymbol{0},\boldsymbol{\sigma}_{\omega_{b }}^2) ; \boldsymbol{\dot{b}}_a \sim \mathcal{N}(\boldsymbol{0},\boldsymbol{\sigma}_{a_b}^2)
\end{equation}

\subsubsection{IMU Propagation}

For two consecutive IMU readouts at time instants $t_{n-1}$ and $t_n$, position, velocity, and orientation states can be propagated by:

\begin{align}
&\boldsymbol{q}_n = \boldsymbol{q}_{n-1} + \frac{1}{2} (\boldsymbol{q}_{n-1} \otimes \boldsymbol{\omega}_n) \Delta t \label{equ:qpropagated}\\
&\boldsymbol{p}_n = \boldsymbol{p}_{n-1} + \boldsymbol{v}_{t-1} \Delta t \\
&\boldsymbol{v}_n = \boldsymbol{v}_{n-1} + (\boldsymbol{R}(\boldsymbol{q}_{n-1})\boldsymbol{a}_n-\boldsymbol{g}) \Delta t \label{equ:vpropagated} \\
&\Delta t= t_n - t_{n-1}
\end{align}
Note that, in the above equations, angular velocity $\boldsymbol{\omega}_n$ and acceleration $\boldsymbol{\omega}_n$ refer to the nominal state, which has been compensated with the corresponding bias ($\boldsymbol{\omega}_n=\boldsymbol{\omega}_{m,n}-\boldsymbol{b}_{\omega}$  and $\boldsymbol{a}_n=\boldsymbol{a}_{m,n}-\boldsymbol{b}_{a}$). The estimation of the biases will be illustrated in subsection \ref{subsection:biasest}. In equation \ref{equ:qpropagated}, $\otimes$ refers to the multiplication operation of the quaternion and angular velocity $\boldsymbol{\omega}=(0, \omega_x, \omega_y, \omega_z)^T$, and it can be calculated by $\boldsymbol{q} \otimes \boldsymbol{\omega} = \Omega (\boldsymbol{\omega}){\boldsymbol{q}}$, where $\Omega (\boldsymbol{\omega})$ is the quaternion integration matrix:
\begin{equation}
\Omega (\boldsymbol{\omega}) = \begin{bmatrix}
0        & -\omega_x & -\omega_y & -\omega_z\\
\omega_x & 0         & \omega_z  & -\omega_y\\ 
\omega_y & -\omega_z & 0         & \omega_x \\ 
\omega_z & \omega_y  & -\omega_x & 0 
\end{bmatrix}
\end{equation}
\subsubsection{Madgwick Feedback}

In order to obtain an attitude estimation of high accuracy, the Madgwick feedback of attitude estimation from the accelerometer to the orientation propagation is adopted. This feedback will be applied when the visual-inertial sensor is close to uniform and exhibits steady motion. When there is no external acceleration (i.e. the sensor is in uniform motion or stays steady), the field of gravity and the field of acceleration measurement should be aligned:

\begin{equation}
\boldsymbol{a}=_w^i\boldsymbol{q}^{*}\otimes\boldsymbol{g}\otimes_w^i\boldsymbol{q}=\boldsymbol{R}(_w^i\boldsymbol{q}) \cdot \boldsymbol{g} = \boldsymbol{R}(\boldsymbol{q}^{*})\cdot\boldsymbol{g}
\end{equation} 
Here, $\boldsymbol{R}(\boldsymbol{q})$ is the rotation matrix and can be calculated by:
\begin{equation}
\boldsymbol{R}(\boldsymbol{q}) =\begin{bmatrix}
1-2q_y^2-2q_z^2&2q_xq_y-2q_zq_w&2q_xq_z+2q_yq_w\\ 
2q_xq_y+2q_zq_w&1-2q_x^2-2q_z^2&2q_yq_z-2q_xq_w\\ 
2q_xq_z-2q_yq_w&2q_yq_z+2q_xq_w&1-2q_x^2-2q_y^2
\end{bmatrix}
\end{equation} 
In this scenario, by aligning the orientations of these two fields, acceleration can be fed back to the orientation estimation. In detail, when the norm of incoming acceleration is close to the magnitude of gravity, i.e. $||\boldsymbol{a}||-||\boldsymbol{g}||< \pi _{sh}$, the orientation estimation from the accelerometer can be presented by:
\begin{equation}
\begin{aligned}
&\boldsymbol{q} = \argmin_{\boldsymbol{q}} \boldsymbol{f}(\boldsymbol{q},\boldsymbol{g},\boldsymbol{a})\\
&\boldsymbol{f}(\boldsymbol{q},\boldsymbol{g},\boldsymbol{a}) = \boldsymbol{R} \left(\boldsymbol{q}^{*} \right ) \cdot \boldsymbol{g}-\boldsymbol{a}
\end{aligned}
\end{equation}
This optimization problem can be solved by the gradient descent algorithm:
\begin{equation}
\boldsymbol{q}_{n+1} = \boldsymbol{q}_{n} - u \frac{ \bigtriangledown \boldsymbol{f}(\boldsymbol{q}_{n},\boldsymbol{g},\boldsymbol{a})}{\parallel \bigtriangledown \boldsymbol{f}(\boldsymbol{q}_{n},\boldsymbol{g},\boldsymbol{a}) \parallel}, n = 0, 1, 2...n
\label{equ:gradientdescent}
\end{equation}
\begin{equation}
\bigtriangledown \boldsymbol{f}(\boldsymbol{q}_{n},\boldsymbol{g},\boldsymbol{a}) = \boldsymbol{J}_f^T(\boldsymbol{q}_{n}) \cdot \boldsymbol{f}(\boldsymbol{q}_{n},\boldsymbol{g},\boldsymbol{a})
\label{equ:gradient}
\end{equation}
Equation \ref{equ:gradientdescent} shows the result of orientation estimation $\boldsymbol{q}_{n+1}$ after $n$ iterations, which is based on an initial guess orientation $\boldsymbol{q}_{0}$ and step size $u$. The gradient of the function can be calculated by its Jacobian and guessed value (Equation \ref{equ:gradient}). For the ENU coordination system, the normalized gravity in the world frame can be presented by $^w\boldsymbol{g}=(0,0,1)^T$. Thus, the objective function $\boldsymbol{f}$ and its Jacobian $\boldsymbol{J}_f^T$ can be derived as follows:

\begin{align}
	&\boldsymbol{f}(\boldsymbol{q}_{n},\boldsymbol{g},\boldsymbol{a}) = \begin{bmatrix} 2q_wq_y-2q_xq+z- a_x \\ -2q_wq_x-2q_yq_z-a_y\\ 2q_x^2+2q_y^2-1-a_z \end{bmatrix}\\
	&\boldsymbol{J}_f^T(\boldsymbol{q}_{n}) = \begin{bmatrix} 2q_y&-2q_z&2q_w&-2q_x\\-2q_x&-2q_w&-2q_z&-2q_y \\0&4q_x&4q_y&0 \end{bmatrix}
\end{align}
Inspired by the work of Madgwick et al. \cite{madgwick2011estimation}, the fusion of orientation estimation from accelerometer to the propagation model can be carried out by fusion of the derivative of orientation $\dot{\boldsymbol{q}}$. In other words, we do not use the $\boldsymbol{q}$ obtained from equation \ref{equ:gradientdescent}. Instead, we take the $\frac{ \bigtriangledown \boldsymbol{f}(\boldsymbol{q},\boldsymbol{g},\boldsymbol{a})}{\parallel \bigtriangledown \boldsymbol{f}(\boldsymbol{q},\boldsymbol{g},\boldsymbol{a}) \parallel}$ part from the first iteration as an approximation of the orientation derivative from the accelerometer $\dot{\boldsymbol{q}}_{a}$. The $\frac{1}{2} \boldsymbol{q}_{n-1} \otimes \boldsymbol{\omega}_n$ term in Equation \ref{equ:qpropagated} is the orientation derivative from the gyroscope. Combining both items with a fusing weighting factor $\pi$, the orientation derivative and fused orientation propagation model can be presented by:

\begin{align}
&\dot{\boldsymbol{q}}=\frac{1}{2}\boldsymbol{q} \otimes \boldsymbol{\omega}+\pi \frac{\bigtriangledown \boldsymbol{f}(\boldsymbol{q}_{n-1},\boldsymbol{g},\boldsymbol{a})}
{\parallel \bigtriangledown \boldsymbol{f}(\boldsymbol{q}_{n-1},\boldsymbol{g},\boldsymbol{a}) \parallel}\\
&\boldsymbol{q}_n = \boldsymbol{q}_{n-1} + \dot{\boldsymbol{q}} \Delta t
\end{align}

\subsection{Visual Estimation with Orientation Feedforward}
The feature-based visual estimation workflow in this research consists of:
\subsubsection{Feature detection and tracking}
Feature detection and tracking: finding the corresponding point pairs between consecutive frames can be carried out either by detection-matching workflow \cite{mur2015orb}, \cite{leutenegger2013keyframe} or by tracking workflow \cite{qin2018vins}. The latter workflow is adopted in consideration of processing speed. Additionally, to ensure robust tracking, several improvements are added to the conventional ORB detection process \cite{rublee2011orb}. The detection starts with extracting all features from the image. Then, all of these features are signed into 16 predefined regions by their distributions in the image plane. In every region, the features are scored with their Harris index \cite{harris1988combined} and sorted accordingly. Followed by the close feature checking, the 15 top score features are selected and added into the feature list. The close feature checking rules out the feature when it is too close to a selected feature.

For the first frame, the feature becomes landmarks after gaining its depth information. For the following frames, after histogram equalization of the image, the features are tracked by the Lucas-Kanade optical flow \cite{bouguet2001pyramidal}. Finally, the tracked point pairs will be verified using the ORB Hamming distance.
\begin{figure}[H]
	\centering
	\begin{subfigure}{0.7\linewidth}
		\centering
		\includegraphics[width=0.95\linewidth]{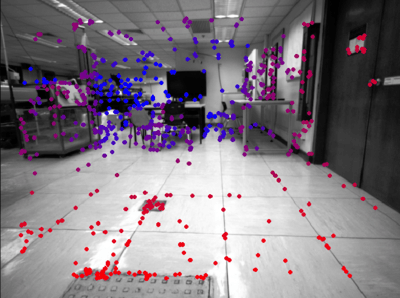}
		\caption{Conventional workflow}
		\label{fig:sfig1}
	\end{subfigure}\\
	\begin{subfigure}{0.7\linewidth}
		\centering
		\includegraphics[width=0.95\linewidth]{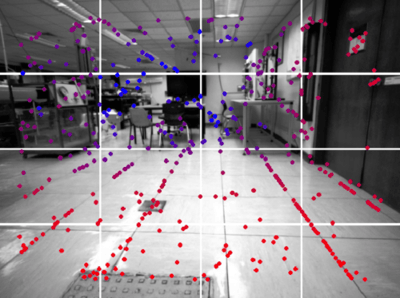}
		\caption{Improved workflow}
		\label{fig:sfig2}
	\end{subfigure}
	\caption{Features of conventional and improved workflows}
	\label{fig:fig}
\end{figure}

\subsubsection{PnP 3D-2D and outlier rejection}
The Perspective-n-Point (PnP) problem refers to the pose estimation of a camera by a set of landmarks and their corresponding 2D projections in the image. For each solution of PnP, the chosen point correspondences cannot be collinear. In addition, PnP can have multiple solutions, and how to choose a particular solution would require postprocessing of the solution set. RANSAC is commonly used with a PnP method to make the solution robust with respect to outliers in the set of point correspondences.
\subsubsection{Roll and pitch feedforward}
As introduced in the previous subsection, with the gradient decrease feedback, the IMU can provide accurate orientation estimation. Moreover, compared to the yaw direction, estimations in pitch and roll directions are more accurate because the gravity components in these two directions are larger than in the z-axis direction. Therefore, we trust the estimation in these two directions and introduce the pitch and roll feedforward prior to the bundle adjustment process. 

\subsubsection{In-Frame Bundle Adjustment} 
This part is the simplified version of sliding window optimization (which will be introduced later in Section \ref{ssec:swo}). The difference is that only the tracked landmarks are considered in the current frame. All landmarks are set as fixed in the optimizer, and the only variable is the camera pose. 

\subsection{Interframe Bias Estimation Feedback}
\label{subsection:biasest}
The biases, which act as errors, accumulate along with the IMU propagation, and they induce drifts in the estimation process. The gyroscope bias induces the orientation drift, while the velocity drift is accumulated by the accelerometer bias and further induces the drift in position. The drift compensation and more accurate egomotion estimation can be achieved by estimating the biases and feeding them back to the IMU propagation model.

\begin{figure}[]
	\centering
	\includegraphics[width=0.8\linewidth]{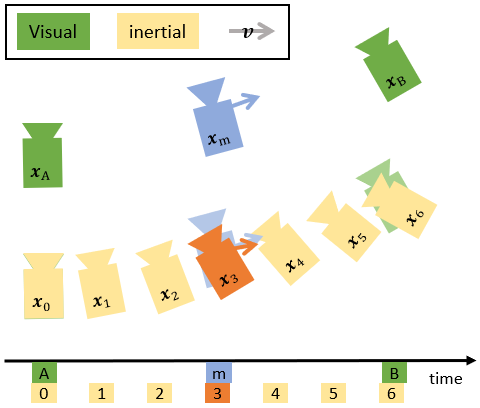}
	\caption{Bias estimation}
	\label{fig:biasest}
\end{figure}

Prior to derivation of bias estimation, some temporarily used notations are introduced. Taking Figure \ref{fig:biasest} as an example, two frames are captured at times $t_A$ and $t_B$. In the interval between these two frames, there exist seven IMU inputs at time $t_0 \cdots t_6$. The relevant vision states and IMU states are presented by $_v\boldsymbol{x}_A$, $_v\boldsymbol{x}_B$ and $_i\boldsymbol{x}_0 \cdots {}_i\boldsymbol{x}_6$. The $\Delta t_A^B$ refers to the time interval between $t_A$ and $t_B$.

The interframe bias estimation is based on three assumptions:
\begin{itemize}
	\item The biases remain constant in the interframe section.
	\item The white noise of the IMU is ignored ($\boldsymbol{n}_{\omega}=\boldsymbol{n}_{a}=0$).
	\item The vision state and IMU state are aligned at the beginning with the same initial state ($_v\boldsymbol{x}_A = {}_i\boldsymbol{x}_0$).
\end{itemize}
Additionally, in real situations, orientation and velocity drifts are coupled. However, in the interframe bias estimation scenario, the coupled term can be neglected because the magnitudes of these drifts are small in a short period of time. Accordingly, the orientation drift and velocity drift are solved separately.

For orientation drift and gyroscope bias estimation, the vision states are considered to be accurate and drift-free and can be propagated by:
\begin{equation}
_v\boldsymbol{q}_n = {}_v\boldsymbol{q}_{n-1}+\frac{1}{2}(_v\boldsymbol{q}_{n-1} \otimes \boldsymbol{\omega}_t)\Delta t_{n-1}^n
\end{equation}
However, biases exist in the IMU propagation and affect the IMU related state. Hence, Equation \ref{equ:qpropagated} becomes:
\begin{equation}
_i\boldsymbol{q}_n = {}_i\boldsymbol{q}_{n-1}+\frac{1}{2}(_i\boldsymbol{q}_{n-1} \otimes (\boldsymbol{\omega}_t+\boldsymbol{b}_{\omega}))\Delta t_{n-1}^n
\end{equation}
As $\Delta t$ is a relatively small value, the above equation can be approximated by IMU states and a bias-induced turbulence:
\begin{equation}
_i\boldsymbol{q}_n \approx {}_i\boldsymbol{q}_{n-1}+\frac{1}{2}(_i\boldsymbol{q}_{n-1} \otimes \boldsymbol{\omega}_t)\Delta t_{n-1}^n  + \frac{1}{2}(_i\boldsymbol{q}_{n-1} \otimes \boldsymbol{b}_{\omega})\Delta t_{n-1}^n
\label{equ:gyrobiasgqin}
\end{equation}
Considering the initial state assumption ($_i\boldsymbol{q}_{n-1} = {}_v\boldsymbol{q}_{n-1}$) and recursive integration over time, Equation \ref{equ:gyrobiasgqin} can be approximated as:
\begin{equation}
\begin{aligned}
_i\boldsymbol{q}_n &\approx {}_v\boldsymbol{q}_n + \frac{1}{2}(_v\boldsymbol{q}_{n-1} \otimes \boldsymbol{b}_\omega)\Delta t_0^n \\
&\approx {}_v\boldsymbol{q}_n + \frac{1}{2}(_v\boldsymbol{q}_{n} \otimes \boldsymbol{b}_\omega)\Delta t_0^n
\end{aligned}
\end{equation}
Multiplying $_v\boldsymbol{q}_n^{-1}$ on both sides yields:
\begin{equation}
_v\boldsymbol{q}_t^{-1} \otimes{} _i\boldsymbol{q}_t \approx 1 + \frac{1}{2}\boldsymbol{b}_\omega\Delta t_0^n
\end{equation}
and the gyroscope bias could be solved as:
\begin{equation}
\boldsymbol{\omega}_b =\frac{2(_v\boldsymbol{q}_t^{-1} \otimes {} _i\boldsymbol{q}_t-1)}{\Delta t_0^n}
\label{equ:gyrobiasest}
\end{equation}
Similarly, for the velocity drift and accelerometer bias, the corresponding vision and IMU states are:

\begin{align}
&_v\boldsymbol{v}_n ={} _v\boldsymbol{v}_{n-1}+(\boldsymbol{R}({}_v\boldsymbol{q}_{n-1})\boldsymbol{a}_n-\boldsymbol{g}) \Delta t_{n-1}^n \\
&_i\boldsymbol{v}_n ={} _i\boldsymbol{v}_{n-1}+(\boldsymbol{R}({}_i\boldsymbol{q}_{n-1})(\boldsymbol{a}_n+\boldsymbol{b}_{a})-\boldsymbol{g}) \Delta t_{n-1}^n
\end{align}
The IMU state could be approximated by:
\begin{equation}
\begin{aligned}
_i\boldsymbol{v}_n &\approx {}_i\boldsymbol{v}_{n-1}+(\boldsymbol{R}({}_i\boldsymbol{q}_{n-1})\boldsymbol{a}_n-\boldsymbol{g}) \Delta t_{n-1}^n
+\boldsymbol{R}({}_i\boldsymbol{q}_{n-1})\boldsymbol{b}_{a} \Delta t_{n-1}^n \\
&\approx {}_v\boldsymbol{v}_{n}+\boldsymbol{R}({}_v\boldsymbol{q}_{n-1})\boldsymbol{b}_{a}  \Delta t_{0}^n \\
&\approx {}_v\boldsymbol{v}_{n}+\boldsymbol{R}({}_v\boldsymbol{q}_{n})\boldsymbol{b}_{a}  \Delta t_{0}^n
\end{aligned}
\end{equation}
Rearranging the above equation and multiplying $\boldsymbol{R}({}_v\boldsymbol{q}_{n}^{-1})$ on both sides yields:
\begin{equation}
\boldsymbol{R}({}_v\boldsymbol{q}_{n}^{-1})(_i\boldsymbol{v}_n-{}_v\boldsymbol{v}_{n}) \approx \boldsymbol{b}_{a}  \Delta t_{0}^n
\end{equation}
and the accelerometer bias could be solved as:
\begin{equation}
\boldsymbol{b}_{a}= \frac{\boldsymbol{R}({}_v\boldsymbol{q}_{n}^{-1})(_i\boldsymbol{v}_n-{}_v\boldsymbol{v}_{n})}{\Delta t_{0}^n}
\label{equ:accbiasest}
\end{equation}
Equations \ref{equ:gyrobiasest} and \ref{equ:accbiasest} are used to estimate the gyroscope bias and accelerometer bias, respectively. As shown in Figure \ref{fig:biasest}, the gyroscope bias is estimated from the derived visual state $\boldsymbol{x}_B$ and the relevant IMU state $\boldsymbol{x}_6$. However, the velocity cannot be derived from a single measurement. Thus, the accelerometer bias is estimated from an interpolated middle state $\boldsymbol{x}_m$ and the corresponding IMU state $\boldsymbol{x}_3$. The middle state contains velocity and orientation parts. The velocity vm can be calculated by the difference of the position ($\boldsymbol{p}$), and the orientation $\boldsymbol{q}_m$ can be derived from the spherical linear interpolation (slerp) operation:

\begin{align}
&\boldsymbol{v}_m = \frac{\boldsymbol{p}_B - \boldsymbol{p}_A}{\Delta t_A^B}\\
&\boldsymbol{q}_m = slerp(\boldsymbol{q}_A,\boldsymbol{q}_B,0.5)\\
\end{align}
The time of the middle state is set to the average of the two frame times, $t_m=(t_A+t_B)/2$.

\subsection{Depth Recovery and IIR Filter}
Some stereo cameras, such as Kinect or D435i, contain the embedded vision processors and can output the rectified depth images. The depth information can be directly read out from the depth image. For conventional stereo camera pairs, a sparse depth recovery is performed. This is used to track the features between stereo images with the Lucas-Kanade optical flow \cite{bouguet2001pyramidal} and then to triangulate tracked point pairs \cite{hartley1997triangulation}.
\begin{figure}[H]
	\centering
	\includegraphics[width=1.0\linewidth]{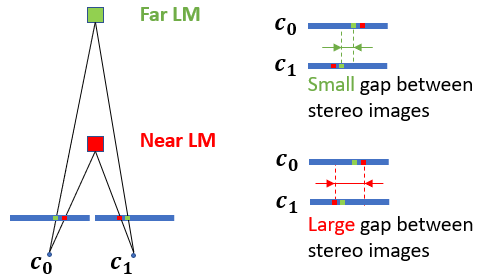}
	\caption{Displacement of near/far landmarks on the image plane. LM: Landmark }
	\label{fig:depthrecoverfail}
\end{figure}
As shown in Figure \ref{fig:depthrecoverfail}, a far landmark and a near landmark are observed by the $c_0$ camera and $c_1$ camera. By aligning the image planes of the two cameras (see the right part of Figure \ref{fig:depthrecoverfail}), it can be observed that the displacement of a certain landmark in the image plane exhibits a negative correlation with its distance from the camera. For the near landmark, there exists a large gap between the corresponding pairs on two image planes. However, because the Lucas-Kanade optical flow can only apply to a limited region on the image plane, the large gap leads to the tracking failure. Therefore, the near landmarks cannot recover their depth information. To address this issue, we improve the workflow with the initial guess and dummy depth technology, as follows:
\begin{itemize}
	\item Reproject landmarks into the $c_1$ frame as an initial guess. For new detected features, the initial guesses are their positions in the $c_0$ frame.
	\item Apply the Lucas-Kanade optical flow from $c_0$ frame to $c_1$ frame.
	\item Triangulate the point pairs. For those untracked features, dummy depths with random depth values are given.
\end{itemize}
Therefore, every feature will gain its depth information from a single capture. For the near landmarks, though their depth information is false in the first loop, there is a high success rate of determining the correct positions in the following loops. After that, the triangulation can be performed correctly.

After the depth recovery, the landmark position is updated by an infinite impulse response (IIR) filter. The inframe landmarks can first be projected into the view of camera $c_0$ (or camera $c_1$ in Figure \ref{fig:depthrecoverfail}):
\begin{equation}
_{c_0}\boldsymbol{lm}_{i-1} = {} _w^{c_0}\boldsymbol{T} _{w} \cdot \boldsymbol{lm}_{i-1}
\end{equation}
The IIR filter is adopted next:
\begin{equation}
_{c_0}\boldsymbol{lm}_{i} = \pi \cdot {} _{c_0}\boldsymbol{lm}_{i-1} + (1-\pi) \cdot {} _{c_0}\boldsymbol{lm}_{i,measure}
\end{equation}
where $\pi$ is the IIR filter’s parameter. A large $\pi$ indicates a long output relay in the historical measurement. According to our experience, the result is not sensitive to the parameter $\pi$.  A typical value of $\pi$ is 0.8. Since every frame can be regarded as an independent measurement, the measurement error follows the Gaussian distribution. Through the IIR filter, the error will converge to zero. Another advantage of adopting the IIR filter is that it utilizes all information throughout the landmark lifespan (from first detection to disappearance).

ex\section{Backend}
The backend of FLVIS is constructed by two parts, local mapping and loop closing. If the motion between the latest frame and last keyframe is larger than a threshold, we will insert the latest frame as a new keyframe and send it to the local mapping thread to update its pose and landmarks. In the loop closing thread, we will attempt to find a loop candidate for the new keyframe and perform pose graph optimization to further refine the corresponding keyframe pose.
\subsection{Local Mapping Thread}
\begin{figure}[H]
	\centering
	\includegraphics[width=0.9\linewidth]{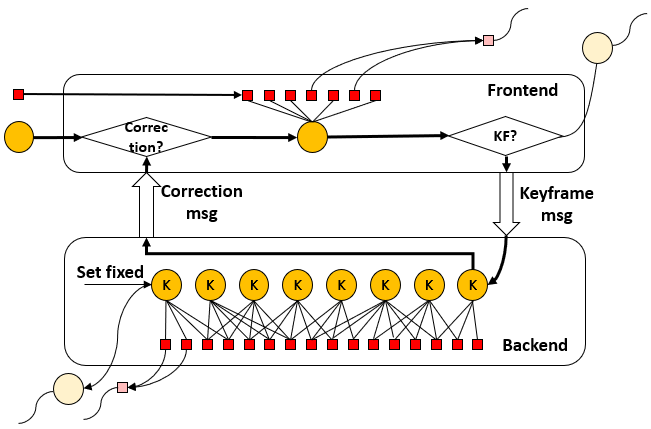}
	\caption{Asynchronous local mapping thread design. KF: Keyframe }
	\label{fig:backend}
\end{figure}
\subsubsection{Coordination with the Frontend using a Selected Keyframe}
The frontend and the backend run asynchronously and are coordinated by the keyframe and correction messages. As shown in Figure \ref{fig:backend}, the frontend tracking thread focuses on the frame to frame tracking, which means that only one frame and its relevant landmarks are kept in the state. Any previous frame and lost tracking landmarks will be immediately excluded. The frontend will publish a keyframe message if one of following criteria is satisfied:
\begin{itemize}
	\item Current frame is the first frame.
	\item The norm of the translation between the current frame and the last keyframe exceeds 0.1 meter.
	\item The norm of the rotation between the current frame and the last keyframe exceeds 0.2 rad.
\end{itemize}
The keyframe message is a snapshot of the current frame, which contains the camera pose and relevant landmarks. At the end of the backend optimization process, a correction message will be sent back to the frontend. Once the frontend notices this correction, it will apply the correction transformation to the pose of the current frame and update the landmark positions accordingly.

\subsubsection{Reprojection Model-Based Sliding Window Optimization}
\label{ssec:swo}
A sliding window with 8 continuous keyframes is maintained. Let $\boldsymbol{m}_{i,j}=(u,v)$ represent the measurement of the landmark (index $i$) in the keyframes (index $j$). The reprojection error is therefore denoted as:
\begin{equation}
\boldsymbol{e}_{i,j} = \boldsymbol{m}_{i,j} - \pi(\boldsymbol{lm}_{i}, {} _c^w\boldsymbol{T}_j) 
\end{equation}
where $\pi$ is the reprojection function. The objective function in the sliding window can thus be presented as:
\begin{equation}
f(\boldsymbol{lm}_{1:n},{}_c^w\boldsymbol{T}_{1:8})= \sum_{i=1}^{n} \sum_{j=1}^{8} \rho_{h}(\boldsymbol{e}_{i,j}^T \Omega_{i,j}^{-1} \boldsymbol{e}_{i,j})
\end{equation}
where $\rho_{h}$ is the Huber robust cost function and $\Omega_{i,j}$ is the covariance matrix which records the measurement relations. $\Omega_{i,j}=\boldsymbol{I}_{2X2}$ indicates a successful observation and $\Omega_{i,j}=\boldsymbol{0}_{2X2}$ denotes that there is no observation of landmark $i$ in frame $j$. The iterative Gaussian Newton method is adopted to minimize the above cost function. In this work, g2o serves as the optimization tool. Note that, in the sliding window, all poses and landmarks except for the oldest pose will be refined. The oldest pose is set as fixed in the optimizer. In total, 20 iterative loops are performed. After 10 loops, those edges which have reprojection errors larger than a threshold are rejected as outliers. The optimizer then refines the remaining inliers for 10 loops as a fine-tuning step.

\subsection{Loop Closure Thread} 
The new keyframe and its pose are then passed to the loop closure thread. To increase the processing speed, only partial detectors and descriptors are extracted in the frontend, which is not sufficient for the loop detection. Therefore, detectors and descriptors are again completely extracted. The loop closure thread is composed of three parts: loop detection, geometry check and loop correction.
\begin{figure}[H]
	\centering
	\includegraphics[width=1.0\linewidth]{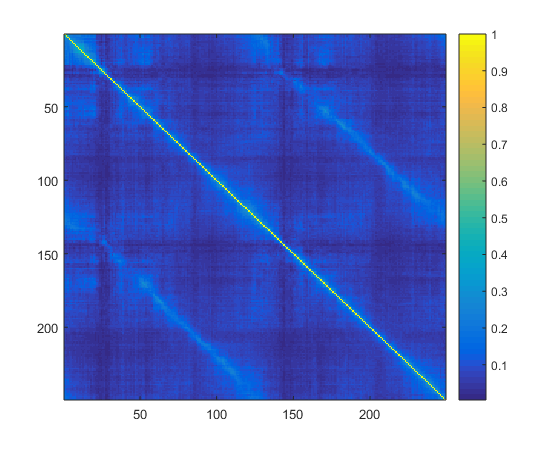}
	\caption{Similarity image of a two-circle trajectory}
	\label{fig:similar}
\end{figure}
\subsubsection{Loop detection}
Similarly to VINS-Mono and ORBSLAM, DBoW (a bag-of-words approach) is applied to find a candidate loop from the keyframe database \cite{GalvezTRO12}. It extracts descriptors from numerous images and then clusters the representative descriptors as words. Descriptors on the queried keyframe will be transferred to a word vector and then compared with previous keyframes by calculating the similarity score.

Figure \ref{fig:similar} shows the similarity image of a two-circle trajectory, where the element in row $i$ and column $j$ indicates the similarity score between keyframes $i$ and $j$. Apparently, a keyframe on the second circle can easily find a loop candidate from the first circle, whose score is the highest, by neglecting the keyframes close to the current keyframe.
Three conditions are set to find the loop candidate:
\begin{itemize}
	\item It has the highest score;
	\item The score is higher than 0.2;
	\item The scores of 3 continuous keyframes before the loop candidate are higher than 0.15.
\end{itemize}

\subsubsection{Geometry check}
If the environment contains similar textures or objects, the wrong loop candidate may be detected. Therefore, a geometry check should be performed to avoid perceptual aliasing. The relative motion between the loop candidate and the current keyframe is calculated by RANSAC PnP. Then, the inlier features are selected by ratio test, cross matching and RANSAC filtering.

The loop candidate passes the geometry test if:
\begin{itemize}
	\item The relative motion is small. Specifically, translation is smaller than 3 m and rotation angle is smaller than 60 degrees.
	\item There exist sufficient matched inliers. Specifically, the number of inliers is larger than 30.
\end{itemize}

\subsubsection{Loop correction}
If the loop candidate passes the geometry test, pose-graph optimization is further conducted to correct the keyframe poses along the loop. This involves two kinds of error: adjacent error and loop error. 

\begin{equation}
\boldsymbol{e}_{m,n} = ln(_c^m\boldsymbol{T}_{n} \cdot _c^w\boldsymbol{T}_{m}^{-1} \cdot _c^w\boldsymbol{T}_{n})
\end{equation}
where, for the adjacent error, $m$ and $n$ represent adjacent keyframe indexes, and for loop error, they represent the end indexes of the loop. The objective function of pose graph optimization is:
\begin{equation}
\sum_{a} \boldsymbol{e}_{a,a+1}^T \Omega_{a,a+1}^{-1} \boldsymbol{e}_{a,a+1} + \sum_{j} \rho_{h}(\boldsymbol{e}_{l1,l2}^T \Omega_{l1,l2}^{-1} \boldsymbol{e}_{l1,l2})
\end{equation}
where $\boldsymbol{e}_{a,a+1}$, $\boldsymbol{e}_{l1,l2}$ and $\Omega$ are the adjacent error, the loop error, and the covariance matrix, respectively. 

Again, the Gaussian Newton method in g2o is applied to minimize the above equation.

\begin{figure*}[!h]
	\centering
	\begin{subfigure}{0.32\linewidth}
		\centering
		\includegraphics[width=0.98\linewidth]{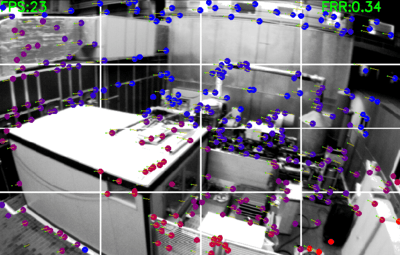}
		\caption{}
	\end{subfigure}
	\begin{subfigure}{0.32\linewidth}
		\centering
		\includegraphics[width=0.98\linewidth]{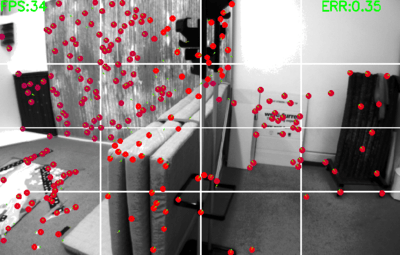}
		\caption{}
	\end{subfigure}
	\begin{subfigure}{0.32\linewidth}
		\centering
		\includegraphics[width=0.98\linewidth]{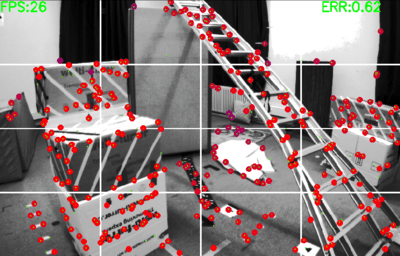}
		\caption{}
	\end{subfigure}\\
	\centering
	\begin{subfigure}{0.32\linewidth}
		\centering
		\includegraphics[width=0.98\linewidth]{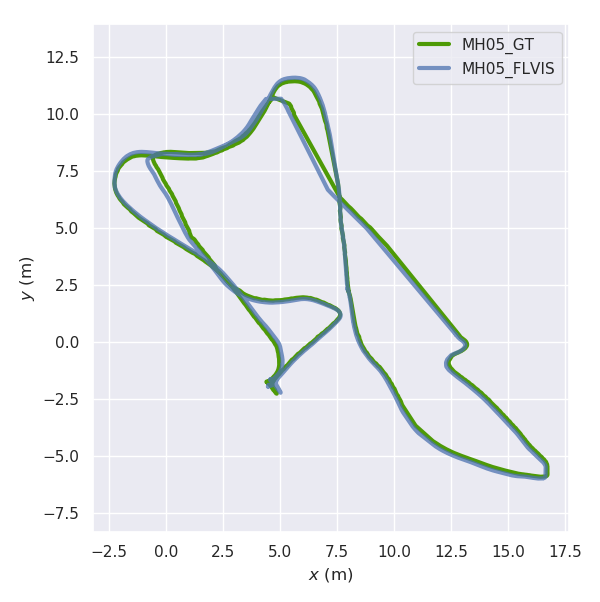}
		\caption{}
	\end{subfigure}
	\begin{subfigure}{0.32\linewidth}
		\centering
		\includegraphics[width=0.98\linewidth]{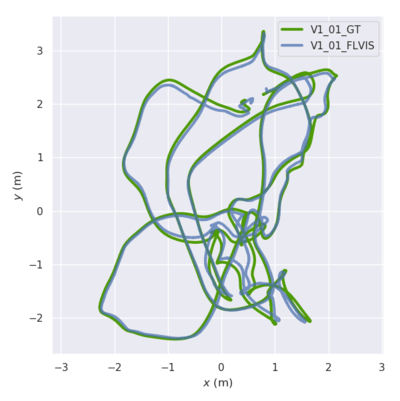}
		\caption{}
	\end{subfigure}
	\begin{subfigure}{0.32\linewidth}
		\centering
		\includegraphics[width=0.98\linewidth]{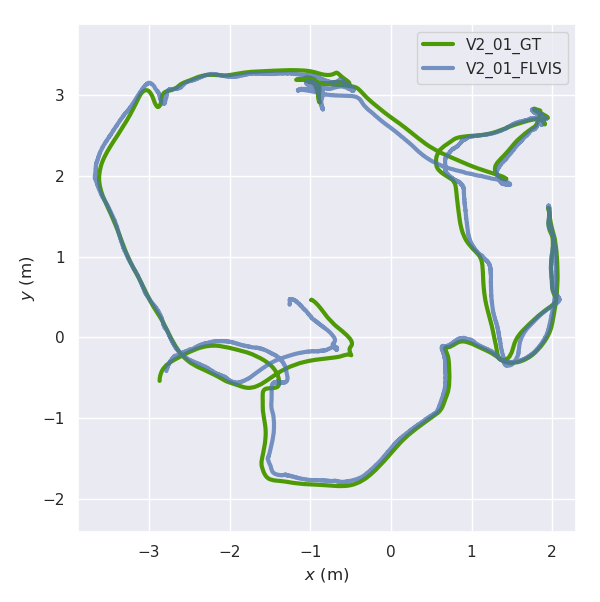}
		\caption{}
	\end{subfigure}\\
	\begin{subfigure}{0.32\linewidth}
		\centering
		\includegraphics[width=0.98\linewidth,height=0.8\linewidth]{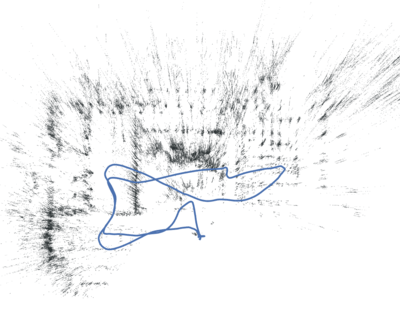}
		\caption{}
	\end{subfigure}
	\begin{subfigure}{0.32\linewidth}
		\centering
		\includegraphics[width=0.98\linewidth,height=0.8\linewidth]{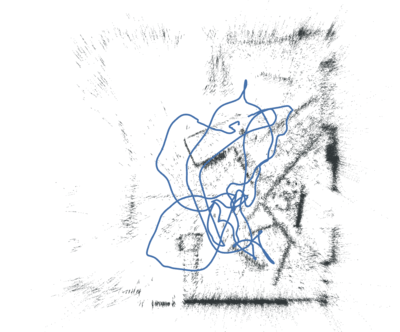}
		\caption{}
	\end{subfigure}
	\begin{subfigure}{0.32\linewidth}
		\centering
		\includegraphics[width=0.98\linewidth,height=0.8\linewidth]{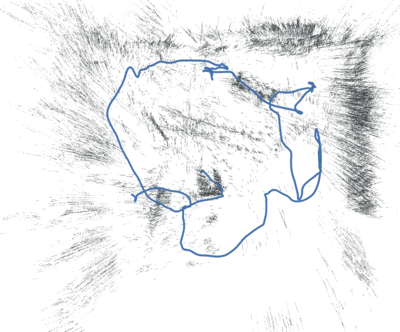}
		\caption{}
	\end{subfigure}
	\caption{FLVIS Results on EuRoC MAV Datasets. (a)-(c) are the scenes of the MH05, V101, and V201 sequences, respectively. (d)-(f) are the comparison of the trajectories (FLVIS) and ground truths (GT), (g)-(i) are the sparse point maps of the sequences; (a)-(d)-(g), (b)-(d)-(h) and (c)-(f)-(i) are the FLVIS results for the corresponding EuRoC MAV sequences.}
	\label{fig:partsofresults}
\end{figure*}

\section{Experimental Results}
We evaluate the proposed method using both the EuRoC MAV Dataset and tests in a real-world environment. In addition, the method is also demonstrated on a resource limited UAV platform. 

The accuracy of the system is presented by the root mean square error (RMSE) of transnational drift. The absolute trajectory error (ATE) is considered \cite{grupp2017evo}. The definitions of ATE and RMSE are shown below:
\textbf{}\begin{align}
&\boldsymbol{E}_{ATE,i}=\boldsymbol{T}_{gt,i}^{-1}\boldsymbol{S}\boldsymbol{T}_{est,i} \\ 
&RMSE(\boldsymbol{E}_{1:n}):=(\frac{1}{n}\sum_{i=1}^{n}\left \| trans(\boldsymbol{E}_{i}) \right \|^{2})^{\frac{1}{2}}
\end{align}
where $\mathbf{T}_{gt,i}$ is the transformation of the ground truth of frame $i$, $\mathbf{T}_{est,i}$ is the transformation estimate of frame $i$ and $\mathbf{S}$ is the least-squares estimation of transformation between the estimated trajectory $\mathbf{T}_{est,1:n}$ and the ground truth trajectory $\mathbf{T}_{gt,1:n}$ by Umeyama’s method \cite{umeyama1991least}.

\subsection{EuRoC Mav Dataset}
The proposed system is first evaluated using the EuRoC MAV Visual-Inertial Datasets \cite{Burri25012016}. These datasets are collected using a UAV mounted VI-sensor, which consists of a stereo module (Aptina MT9V034 global shutter, WVGA monochrome, 20 FPS) and a hardware synchronized MEMS IMU (ADIS16448, angular rate and acceleration, 200 Hz). The ground truths of these datasets are provided by a VICON (6D pose, Vicon Room sequences) and a Leica MS50 (3D position, Machine Hall sequence). The datasets are run on a standard laptop (CPU: Core i5-7200U; RAM: 8GB).

For the comparison group, the notable work of the external Kalman filter-based (MSCKF\footnote{https://github.com/KumarRobotics/msckf\_vio} \cite{mourikis2007multi} \cite{sun2018robust}) and optimization-based (VINS-Fusion\footnote{https://github.com/HKUST-Aerial-Robotics/VINS-Fusion} \cite{qin2019general}) methods are selected. All algorithms are run in the stereo-visual-inertial mode and disable the loop closure function. Selected results are listed in Figure \ref{fig:partsofresults}, and the ATEs for all sequences are listed in Table \ref{tab:comparison_ate}.
\begin{table}[H]
	\centering
	\caption{ATE Comparison in EuRoC MAV Datasets}
	\begin{tabular}{|l|c|c|c|c|}
		\hline
		\multicolumn{1}{|c|}{\multirow{2}{*}{Sequence}} & \multirow{2}{*}{Length} & \multicolumn{3}{c|}{ATE}         \\ \cline{3-5} 
		\multicolumn{1}{|c|}{}                          &                         & FLVIS & VINS-Fusion & MSCKF \\ \hline
		MH 01   easy                                    & 80.96                   & \textbf{0.15}       & 0.27                 & -              \\ \hline
		MH 02   easy                                    & 73.22                   & \textbf{0.11}       & 0.20                 & 0.17           \\ \hline
		MH 03 medium                                    & 127.07                  & 0.31                & 0.35                 & \textbf{0.25}  \\ \hline
		MH 04 difficult                                 & 91.10                   & \textbf{0.25}       & 0.44                 & 0.51           \\ \hline
		MH 05 difficult                                 & 96.98                   & \textbf{0.28}       & 0.37                 & 0.35           \\ \hline
		V1 01    easy                                   & 60.12                   & \textbf{0.10}       & 0.13                 & \textbf{0.10}  \\ \hline
		V1 02  medium                                   & 74.68                   & 0.15                & \textbf{0.14}        & 0.16           \\ \hline
		V1 03  difficult                                & 75.10                   & 0.24                & \textbf{0.13}        & 0.20           \\ \hline
		V2 01    easy                                   & 36.58                   & 0.17                & 0.14                 & \textbf{0.09}  \\ \hline
		V2 02  medium                                   & 81.47                   & 0.30                & \textbf{0.20}        & 0.23           \\ \hline
		V2 03  difficult                                & 83.52                   & -                   & \textbf{0.39}        & -              \\ \hline
	\end{tabular}
	\vspace{1ex}
	
	{\raggedright Note: MH: Machine Hall sequence; V: Vicon Room sequence \par}
	\label{tab:comparison_ate}
\end{table}

As shown in Table \ref{tab:comparison_ate}, all approaches achieve high accuracy (i.e. $ATE/Length<1\%$). The current FLVIS approach outperforms or performs equally to the other two with respect to most Machine Hall sequences and one Vicon Room sequence. Since every stereo image is considered with equal variability in our FLVIS method, the nonuniform exposure of the stereo images affects the accuracy of the system (such as in the V1 03 and V2 03 sequences). As shown in Figure \ref{fig:partsofresults}, good agreement between the estimated trajectory and the ground truth can be observed. The sparse point map can reflect the environment settings in Machine Hall and Vicon Room.

\subsection{Real-World Environment with D435i sensors}
The real-world test using the RealSense D435i sensor was conducted. The sensor contains two global shutter cameras (resolution: 640x480, run at 30 Hz), an IMU (run at 450 Hz) and an integrated vision processing unit which can directly output the depth image. In contrast to the previous test, the algorithm here directly uses the depth output of the camera instead of triangulation from the stereo images, leading to reduced processing time. The real-world test contains 2 sequences (i.e. with/without loop closure thread enabled), which cover some challenging scenarios (Figure 10). The complete video clips can be found in the supplemental material.

\begin{figure*}[!h]
	\centering
	\begin{subfigure}{0.24\linewidth}
		\centering
		\includegraphics[width=0.98\linewidth,height=0.8\linewidth]{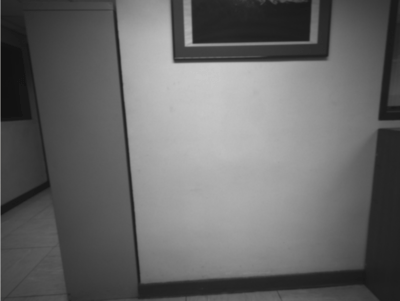}
		\caption{}
	\end{subfigure}
	\begin{subfigure}{0.24\linewidth}
		\centering
		\includegraphics[width=0.98\linewidth,height=0.8\linewidth]{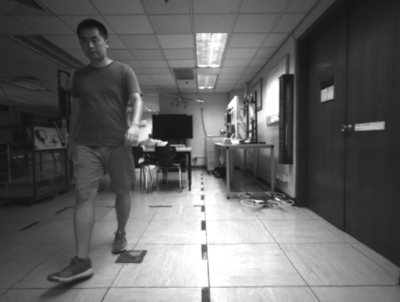}
		\caption{}
	\end{subfigure}
	\begin{subfigure}{0.24\linewidth}
		\centering
		\includegraphics[width=0.98\linewidth,height=0.8\linewidth]{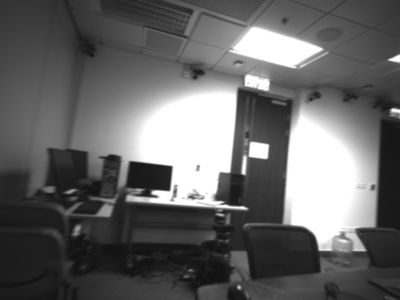}
		\caption{}
	\end{subfigure}
	\begin{subfigure}{0.24\linewidth}
		\centering
		\includegraphics[width=0.98\linewidth,height=0.8\linewidth]{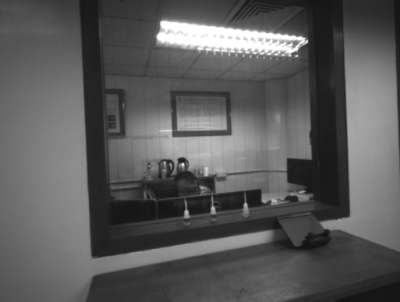}
		\caption{}
	\end{subfigure}
	\caption{Images with (a) poorly-detailed features, (b) pedestrian, (c) large overexposure area, and (d) glass window, presenting some challenging scenarios in the real-world environment.}
	\label{fig:challenge}
\end{figure*}

\begin{figure}[H]
	\centering
	\includegraphics[width=0.8\linewidth]{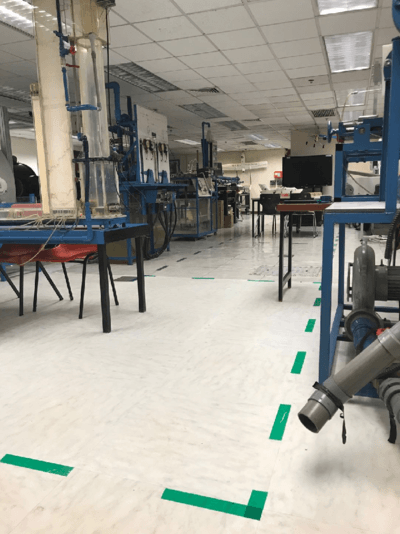}
	\caption{Lab Environment} 
	\label{fig:lavenvironment}
\end{figure}

\begin{figure}[]
	\centering
	\begin{subfigure}{0.48\linewidth}
		\centering
		\includegraphics[width=0.98\linewidth]{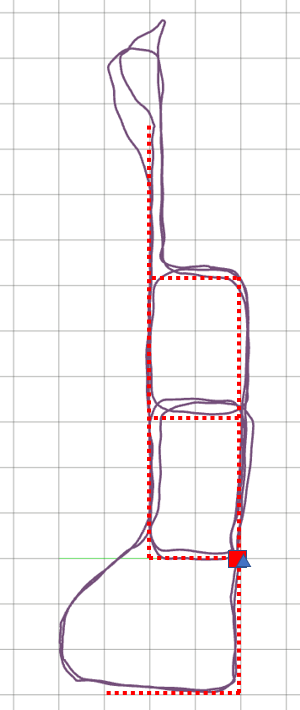}
		\caption{Without loop closure} 
		\label{fig:withoutlc}
	\end{subfigure}
	\begin{subfigure}{0.48\linewidth}
		\centering
		\includegraphics[width=0.98\linewidth]{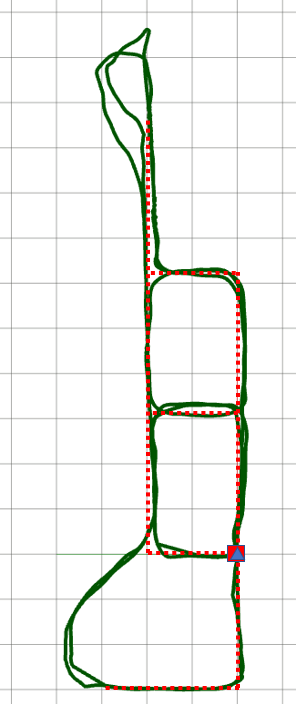}
		\caption{With loop closure}
		\label{fig:withlc}
	\end{subfigure}
	\caption{Trajectories with/without loop closure in the laboratory scenario. The start and end points of the trajectories are marked as red rectangle and blue triangle; the grid size of the figure is 1mx1m.} 
	\label{fig:withwithoutlc}
\end{figure}

Figure \ref{fig:withwithoutlc} shows the estimated trajectories in our laboratory with and without the loop closure thread enabled. The red dotted line along the grid in Figure \ref{fig:withwithoutlc} represents the designed walking trajectory marked by the dark tape on the floor of the laboratory (Figure \ref{fig:lavenvironment}). In this test, the experimentalist starts at the red mark, walks along the designed trajectory in the laboratory for several turns and finally returns to where the test started. When the experimentalist passes the same place for the second time, the loop closure thread identifies the scenario and optimizes the loop poses between two passes. As shown in the figure, for the no loop closure trajectory, there exist certain observable drifts, causing it to be unable to return to the starting point. In contrast, the trajectory with loop closure is much more consistent than that without loop closure. 

\subsection{Typical Processing Time of FLVIS in Different Platforms} 
The FLVIS algorithm is tested in several platforms, varying from an embedded x5-z8350 processor to a powerful i7-8665U platform. The processing times are measured on the basis of MH 05 sequences in the EuRoC dataset (stereo mode) and the real-world lab sequences (depth mode). The average processing time for each part of the algorithm is listed in Table \ref{tab:processingtime}. The processing time of the backend is affected by the size of the sliding window. However, this will not affect the frontend processing time. Meanwhile, the processing time of FLVIS is compared with the conventional optimization-based system (VINS-Fusion). As shown in Table II, FLVIS has greater FPS (frame per second) and runs faster than VINS-Fusion. In addition, with the VPU integrated sensor, which can output the depth directly, FLVIS can further achieve higher FPS. FLVIS is demonstrated to be suitable for the resource limited robotic system. 

\begin{table}[H]
	\caption{Processing Time of Each Part of the Algorithm}
	\begin{tabular}{|c|c|c|c|c|c|}
		\hline
		\multicolumn{2}{|c|}{}                            & C1    & C2    & C3    & C4    \\ \hline
		\multirow{5}{*}{Frontend} & tracking (ms)         & 13.7  & 10.7  & 6.6   & 2.7   \\ \cline{2-6} 
		& pose estimation (ms)       & 3.8   & 2.5   & 1.8   & 1.1   \\ \cline{2-6} 
		& feature detection (ms)        & 43.7  & 24.0  & 14.6  & 7.1   \\ \cline{2-6} 
		& depth recovery-stereo (ms) & 25.3  & 17.0  & 18.3  & 3.9   \\ \cline{2-6} 
		& depth recovery-depth (ms)  & 1     & 1     & 1     & 1     \\ \hline
		\multirow{3}{*}{Backend}  
		& 4 keyframes (ms)           & 53    & 22.0  & 13.7  & 9.8   \\ \cline{2-6} 
		& 8 keyframes (ms)          & 172   & 63.8  & 34.2  & 21.1  \\ \hline
		\multicolumn{2}{|c|}{FPS FLVIS stereo/depth}     & 11/16 & 15/20 & 24/42 & 67/83 \\ \hline
		\multicolumn{2}{|c|}{FPS VINS-Fusion stereo}      & 9     & 13    & 21    & 41    \\ \hline
	\end{tabular}
	\vspace{1ex}
	
	{\raggedright Note: C1: UP-board (x5-z8350, 4G RAM); C2: LattePanda 864s (m3-8100Y, 8G RAM); C3: Laptop (i5-7200U, 8G RAM); C4: Intel NUC (i7-8665U, 16G RAM) \par}
	\label{tab:processingtime}
\end{table}

\subsection{Test on UAV}
\begin{figure}[H]
	\centering
	\includegraphics[width=0.9\linewidth]{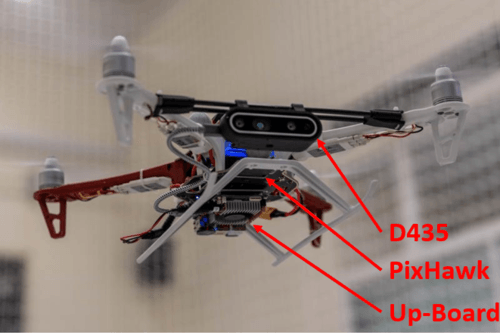}
	\caption{The UAV experimental platform}
	\label{fig:uav_mount}
\end{figure}
\begin{figure}[H]
	\centering
	\includegraphics[width=0.9\linewidth]{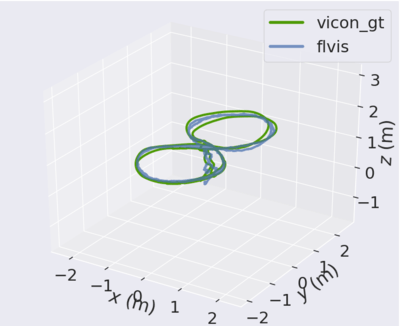}
	\caption{The estimated UAV trajectory and the ground truth}
	\label{fig:uav_trj}
\end{figure}
FLVIS is applied on a UAV platform to achieve autonomous feedback control in a VICON motion capture laboratory. Notably, the loop closure thread is not enabled in the UAV test. As shown in Figure \ref{fig:uav_mount}, the UAV platform consists of a forward looking depth camera (D435i), an IMU integrated fly controller (Pixhawk) and an on-board computer (Up-Boaad). The UAV is programmed to follow the figure-8 trajectory with the radius of 1 meter for each loop. The ground truth is captured by the VICON system. The results exhibit good agreement between the estimated trajectory and ground truth (Figure \ref{fig:uav_trj}). The total length of the trajectory is 31.0 meters, and the ATE is 0.2 meters. The full video of this experiment can be found in the supplemental materials.

\section{Conclusion}
In this paper, we propose a new feedback loops-based stereo visual-inertial framework, named FLVIS (Feedforward-feedback Loop-based Visual Inertial System). The new FLVIS replaces the conventional fusion models (such as extened Kalman filter) by adopting a new control-based strategy. In FLVIS, the objective of fusion is no longer to determine the most appropriate pose guess to satisfy every single measurement (visual and IMU) in a certain period of time. Instead, FLVIS provides a good initial pose value and maintains more reliable landmark positions. The accuracy of FLVIS is secured by the reprojection model-based optimization. Moreover, the frontend and backend are further decoupled with the keyframe-correction mechanism. Covisibility-based bundle adjustments are moved to the backend so that the frontend is kept simple and fast. We then compare FLVIS with other state-of-the-art open source implementations on the public datasets. The superior performance of FLVIS is presented. Furthermore, FLIVS is demonstrated in a resource limited UAV platform. The implementation of FLVIS is open source for the benefit of the viSLAM community.

\section*{Acknowledgments}
This research is supported by EMSD HongKong under Grant. DTD/M\&V/W0084/S0016/0523. The authors would like to thank Jeremy Chang and Weifeng Zhou from MAV LAB, PolyU for their support with hardware, video recording and experimentation.

\ifCLASSOPTIONcaptionsoff
  \newpage
\fi



\bibliographystyle{IEEEtran}
\bibliography{root}

\vfill


\end{document}